\documentclass[letterpaper, 10pt, conference]{ieeeconf}      

\IEEEoverridecommandlockouts                              

\title{ \LARGE \bf
TMSTC*: A Turn-minimizing Algorithm For Multi-robot Coverage Path Planning 
}

\usepackage{amsmath}
\usepackage{amsmath,amssymb}
\usepackage[ruled]{algorithm2e}
\usepackage{courier}
\usepackage{graphicx} 
\usepackage{float} 
\usepackage{subfigure} 
\usepackage{booktabs}
\usepackage[font=small]{caption}
\usepackage{multirow}
\usepackage{multicol}
\usepackage{array}
\usepackage[colorlinks,linkcolor=blue]{hyperref}

\author{Junjie Lu, Bi Zeng$^{\dag}$, Jingtao Tang, and Tin Lun Lam
	\thanks{Junjie Lu and Bi Zeng are with School of Computer Science and Technology, Guangdong University of Technology, Guangzhou (email: \href{courierlo@163.com}{courierlo@163.com}; \href{zb9215@gdut.edu.cn}{zb9215@gdut.edu.cn}).}
	\thanks{Jingtao Tang and Tin Lun Lam are with School of Science and Engineering, The Chinese University of Hong Kong, Shenzhen (email: \href{todd.j.tang@gmail.com}{todd.j.tang@gmail.com}; \href{tllam@cuhk.edu.cn}{tllam@cuhk.edu.cn}).}
	\thanks{$^{\dag}$Corresponding author}
}

\begin{document}

\maketitle
\thispagestyle{empty}
\pagestyle{empty}

\begin{abstract}

Coverage path planning is a major application for mobile robots, which requires robots to move along a planned path to cover the entire map. For large-scale tasks, coverage path planning benefits greatly from multiple robots. In this paper, we describe Turn-minimizing Multirobot Spanning Tree Coverage Star(TMSTC*), an improved multirobot coverage path planning (mCPP) algorithm based on the MSTC*. Our algorithm partitions the map into minimum bricks as tree’s branches and thereby transforms the problem into finding the maximum independent set of bipartite graph. We then connect bricks with greedy strategy to form a tree, aiming to reduce the number of turns of corresponding circumnavigating coverage path. Our experimental results show that our approach enables multiple robots to make fewer turns and thus complete terrain coverage tasks faster than other popular algorithms.
\end{abstract}

\section{Introduction}

Coverage path planning (CPP) refers to the use of mobile robots that traverse the target environment area within their physical contact range or within their sensor sensing range, and the planning performance are generally measured by three metrics: coverage time, repeat coverage area size, and coverage rate \cite{galceran2013survey}. Applications of coverage path planning appear in many aspects of military, agriculture, industry, commerce, disaster relief, and urban life. Substantive instances include automated cleaning \cite{yasutomi1988cleaning}, crop harvesting \cite{hameed2014intelligent}, aerial robot patroling \cite{acevedo2013distributed}, and robotic demining \cite{acar2003path}. In sight of the wide range of applications, research related to coverage path planning has received extensive attention.

In general, coverage tasks have a more pronounced spatial parallelism and can be processed in parallel. Therefore, with the increasing size of overlays and the development of multi-robot technology, among other factors, multi-robot systems have been introduced to overlays with the expectation of acceleration and thus achieving better benefits. The use of multiple robots has several advantages over single-robot systems. First, collaborating robots have the potential to complete tasks faster because the task can be decomposed into sub tasks and distributed to robots that execute the sub tasks in parallel. Second, by employing several robots, redundancy is significantly increased, so teams of robots have better fault tolerance than just one robot. Another advantage of team robots is the incorporation of sensor information that can compensate for sensor uncertainty. Multiple robots have been shown to be more efficient at self-localization, especially when they have different sensors. However, using team robots to cover areas brings about the challenges of coverage algorithms in robot communication, localization, collaboration, and conflict resolution. Many approaches extend single-robot algorithms to multiple robots by decomposing and distributing the workload \cite{almadhoun2019survey}.

The robot is often assumed to move at constant speed in coverage mission, so the path length is used as the only cost. However, in the reality the robot needs to decelerate during turns and as the number of turns increases, the time to complete the coverage task also has to increase. Therefore, for algorithms based on spanning tree coverage, the structure of the tree can have a crucial impact on the coverage time. A better strategy is to cut down the number of turns while reducing the path length. Nevertheless, coverage planning with minimum turns is an NP-Hard problem \cite{arkin2000approximation}, \cite{arkin2005optimal}. In this article, we look for spanning trees with ideal shapes that allow multiple robots to make fewer turns when performing spanning tree coverage, thus saving the completion time of the coverage task. We propose a new method called TMSTC* (Turn-minimizing Multirobot Spanning Tree Coverage Star), which is an extension to MSTC* \cite{tang2021mstc}, to reduce the number of turns for multirobot coverage and to minimize the completion time by averaging the path cost of each robot with the number of turns. We summarize the main contributions of this paper below:

\begin{enumerate}[]
	\item  We propose an effective algorithm to partition the map with a minimum number of bricks as spanning tree's branches.
	\item  We design a greedy algorithm to connect bricks to form a complete spanning tree and then apply the greedy strategy of MSTC* to find the optimal equilibrium division on the topological loops around the spanning tree, thus averaging the weights of each robot coverage path. In addition, we have incorporated turn cost into the weights of the paths.
	\item We present extensive simulation in different kinds of environments. The results show that our method outperforms other popular STC-based algorithms and the state-of-the-art method from \cite{vandermeulen2019turn} in minimizing the number of turns and the coverage time.
	\item We provide an open-source ros-package of our algorithm at \href{https://github.com/CourierLo/TMSTC-Star}{https://github.com/CourierLo/TMSTC-Star}.
\end{enumerate}

\section{Related Work}
\begin{figure}[!htbp]
	\vspace{3pt}
	\centering
	\includegraphics[width=0.4\textwidth]{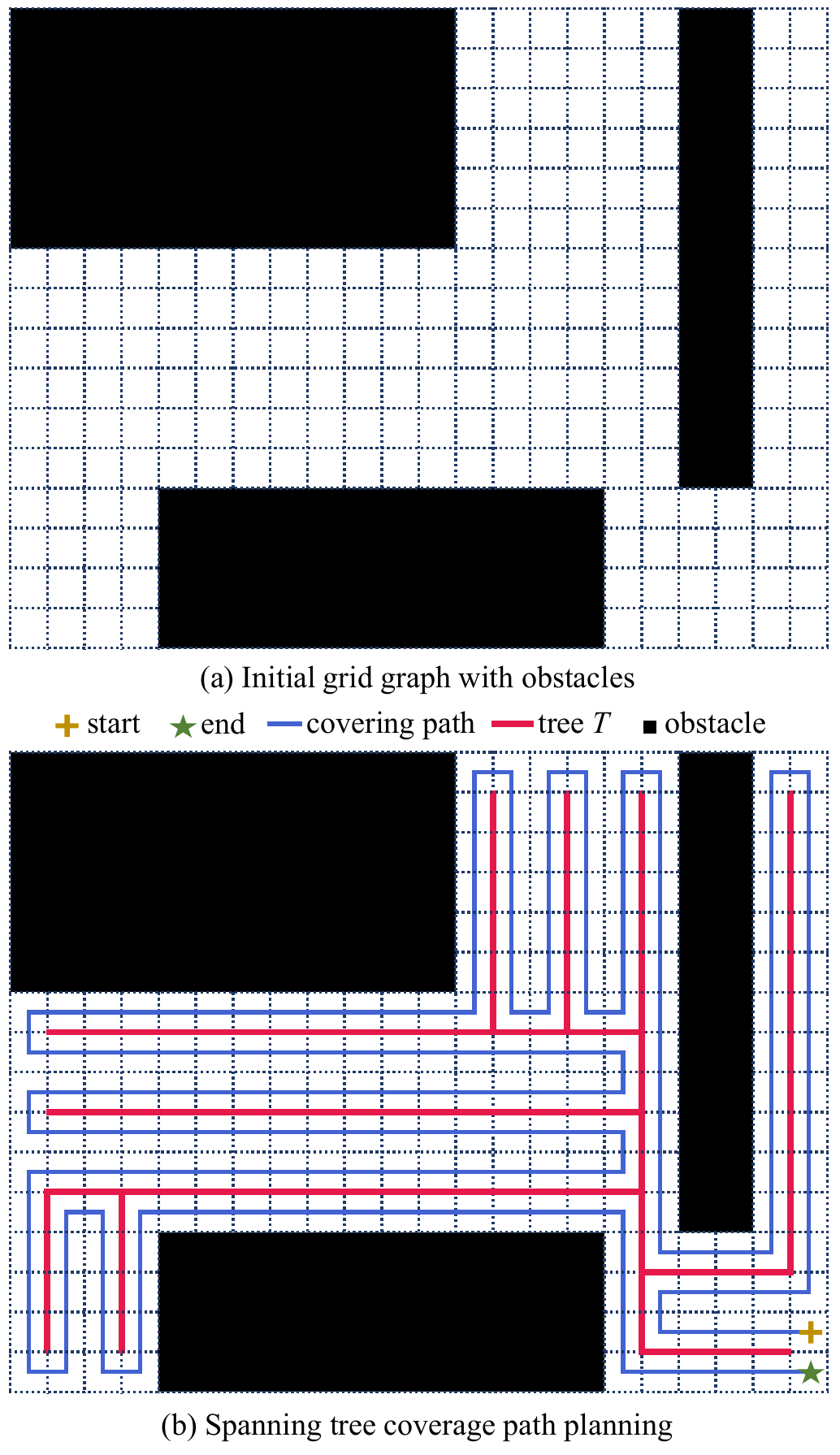}
	\caption{Example of single robot spanning tree coverage path planning. The initial map is divided by $2\times2$ mega cells and every mega cell consists of four $1\times1$ cells. The robot will move counterclockwise along the spanning tree until it has reached each non-occupied grid in known terrain. The coverage path of the robot is a loop.}
	\label{Fig.STC}
	\vspace{-1em}
\end{figure}

A fundamental method that has attracted much attention is spiral spanning tree coverage (STC) for single robot proposed by Gabreiy \cite{gabriely2001spanning}, \cite{gabriely2002spiral}. An example of single robot spanning tree coverage algorithm is shown in Fig. \ref{Fig.STC}. Hazon et al. extended STC to multiple robots and proposed the multi-robot spanning tree coverage (MSTC) with backtracking optimization, which improved the efficiency and robustness of robot coverage \cite{hazon2005redundancy,hazon2006towards}. Agmon et al. extended the MSTC algorithm by varying the shape of the spanning tree to average the workload of each robot \cite{agmon2006constructing}. Zheng et al. proposed the MFC algorithm, a new polynomial-time multirobot coverage algorithm based on an algorithm for finding a tree cover with trees of balanced weights \cite{even2004min}. The algorithm works for both unweighted terrains \cite{zheng2005multi} and weighted terrains \cite{zheng2007robot,zheng2010multirobot}. Tang et al. considered physical constraints such as charging and reloading, improved the MSTC algorithm and proposed MSTC*, which uses a greedy strategy to enable load balancing of multiple robots\cite{tang2021mstc}. Sun et al. proposed a new efficient fault tolerance algorithm, which is an extension of MSTC* \cite{sun2021ft}. The DARP algorithm proposed by Kapoutsis et al. divides the terrain into a number of equal regions, each corresponding to a specific robot, and each robot applies STC algorithm to finish coverage task \cite{kapoutsis2017darp}. Apostolidis et al. presented a high level, user-friendly platform developed for performing multi-UAV coverage missions using DARP \cite{apostolidis2022cooperative}. Gao et al. used the ant colony optimization to optimize the shape of the spanning tree and combined it with the DARP algorithm \cite{gao2018optimal}. Vandermeulen et al. proposed the state-of-the-art coverage method with multiple robots using iterative heuristic algorithm to minimize the number of coverage lines so as to reduce the number of turns in the path, and they used an $m$-TSP algorithm to find paths for each robot \cite{vandermeulen2019turn}. Ramesh et al. developed the OARP coverage planning method that aims to minimize axis-parallel coverage lines for single robot coverage task by using a linear program \cite{ramesh2022optimal}. They formulated the problem into a mixed integer linear program (MILP) and then proved that the linear relaxation provides optimal solutions.

\section{Problem Definition \& Preliminaries}

\subsection{Notation} 
\begin{figure}[htbp]
	\vspace{0em}
	\centering
	\includegraphics[width=0.45\textwidth]{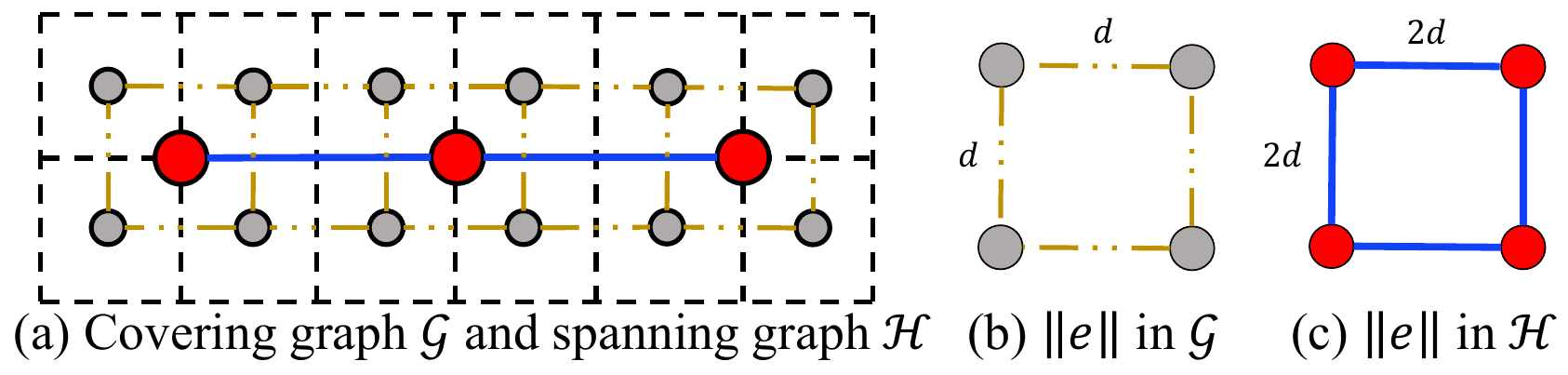}
	\captionsetup[figure]{font=small}
	\caption{Covering graph $\mathcal{G}$ is represented by grey nodes and dotted edges, while spanning graph $\mathcal{H}$ is represented by red nodes and solid edges. A spanning node is generated from 4 adjacent covering nodes. The length of the edges of $\mathcal{G}$ and $\mathcal{H}$ are $d$ and $2d$, respectively.}
	\label{Fig.Graph}
	\vspace{-1em}
\end{figure}

The goal of mCPP is to cover all locations in known terrain using multiple ($k$) robots. Here, we focus on robots that operate on flat surfaces, such as cleaning robots. Assuming that the robot covering tool has a width of $d$, we discretize the terrain to be covered into a large square grid, each with a width of $2d$, containing four small squares of width $d$. Therefore, we obtain two graphs, the coverage graph $\mathcal{G}$ and the spanning graph $\mathcal{H}$. The nodes of two graphs are coverage nodes $\pi$ and spanning nodes $\rho$, respectively. The form of two graphs and corresponding edge weights $\Vert e \Vert$ are shown in Fig. \ref{Fig.Graph}.

\subsection{mCPP Problem Definition}
Given $\mathcal{G}$, $\mathcal{H}$ and $k$ robots, we let the set of points through which robot $R_i$ passes be \{$\pi_i$\} and these points are from path $\Pi_i$. The cost of the path is denoted by $\mathcal{W}_{\Pi_i}$. Ideally, each robot has the same workload and completes its respective coverage tasks at the same time. Therefore, we aim at finding a set of coverage path \{$\Pi_1$, $\Pi_2$ , $\cdots$ , $\Pi_k$\} for $k$ robots while minimizing the maximum of $\mathcal{W}_{\Pi_i}$. That is

\begin{equation}
	\underset{\left\{\Pi_{i}\right\}}{\arg \min }\left(\max _{1 \leq i \leq k}\left(\mathcal{W}_{\Pi_{i}}\right)\right)
\end{equation}

\subsection{Turn-minimizing mCPP Pipeline}
\begin{figure}[htbp]
	\vspace{0em}
	\centering
	\includegraphics[width=0.45\textwidth]{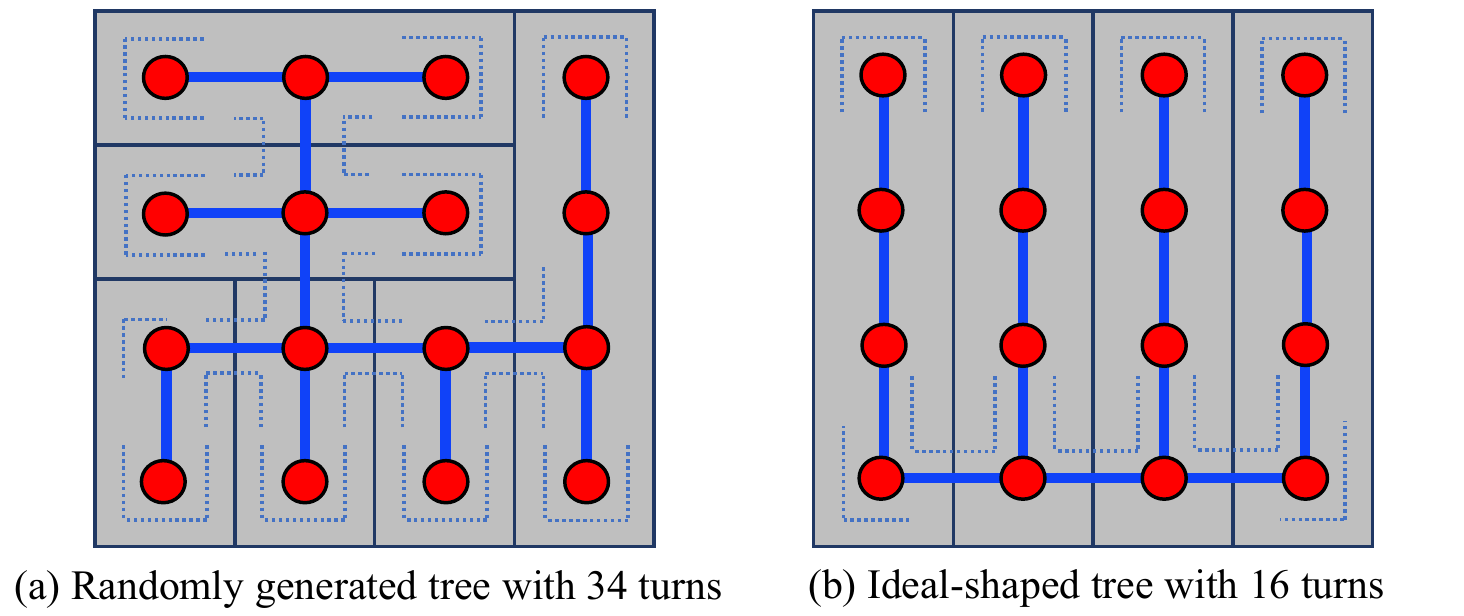}
	\caption{Comparison of the number of turns in different spanning trees. Each gray rectangle represents a brick. There are 6 bricks in (a) and 4 bricks in (b). A well constructed spanning tree substantially reduces the number of turns.}
	\label{Fig.STC_comp}
	\vspace{-1em}
\end{figure}

Fig. \ref{Fig.STC_comp} shows us that the shape of spanning tree will greatly affect the number of turns in robots path. Randomly generated spanning tree increases the number of turns in coverage task dramatically. Turns can be reduced by constructing spanning tree with long straight bricks(see Fig. \ref{Fig.STC_comp}-(b)). A brick $B$ is defined as a rectangle with integer side length, i.e., with width of $2d$ or height of $2d$ (or both) in $\mathcal{H}$. A brick is also considered as a sequence of grids in $\mathcal{H}$ with the $(x, y)$-coordinates

\begin{equation}
	B_i = (\rho_{1}^i, \rho_{2}^i, \cdots, \rho_{n}^i)
\end{equation}
where $x_{\rho_{1}^i} = x_{\rho_{2}^i} = \cdots = x_{\rho_{n}^i}$ or $y_{\rho_{1}^i} = y_{\rho_{2}^i} = \cdots = y_{\rho_{n}^i}$, $\rho_{j}^i$ and $\rho_{j+1}^{i}$ are adjacent. $\rho_{j}^i$ denotes the $j$-th element in $B_i$.

We aim to tile $\mathcal{H}$ with minimum bricks, such that each non-occupied grid is covered by exactly one brick, and each occupied grid is not covered by any brick. Bricks do not overlap. Since the robots are only required to turn between two connecting or adjacent bricks, we can minimize the turns by minimizing the number of bricks. The minimum brick tiling is represented as
\begin{equation}
\begin{aligned}
	& \min{ \left\| \{ \mathnormal{B_1, B_2, \cdots, B_m} \} \right\| }\\
	& \begin{array}{r@{\quad}l@{}l@{\quad}l}
	s.t. &B_{1} \cup B_{2} \cup \cdots \cup B_m = \mathcal{H} \\
	&B_i \cap B_j = \varnothing, \quad \forall i, j \in [1,m], i \ne j
\end{array}
\end{aligned}
\end{equation}

After merging those bricks into a spanning tree with greedy strategy, we use improved MSTC* algorithm to plan a set of coverage paths \{$\Pi_i$\} for $k$ robots.

In short, our turn-minimizing algorithm for mCPP consists of three stages.
\begin{enumerate}[]
	\item Partitioning spanning graph $\mathcal{H}$ with minimum non-overlap bricks
	\item Merging bricks with greedy strategy for further reduction of turns
	\item Computing paths \{$\Pi_i$\} for robots with improved MSTC* algorithm
\end{enumerate}

\section{Our Algorithm}

\subsection{Minimum Brick Tiling In Spanning Graph}
It is not easy to find a minimum brick tiling in a grid graph directly. Noticing that each pair of adjacent grids in spanning graph is divided by a line segment of length $2d$, we choose to delete line borders between two adjacent non-occupied grids as many as possible so as to form a minimum brick tiling. However, a brick must not be L-shaped because bricks are rectangles as we have mentioned before. 

\begin{figure}[htbp]
	\vspace{0em}
	\centering
	\includegraphics[width=0.45\textwidth]{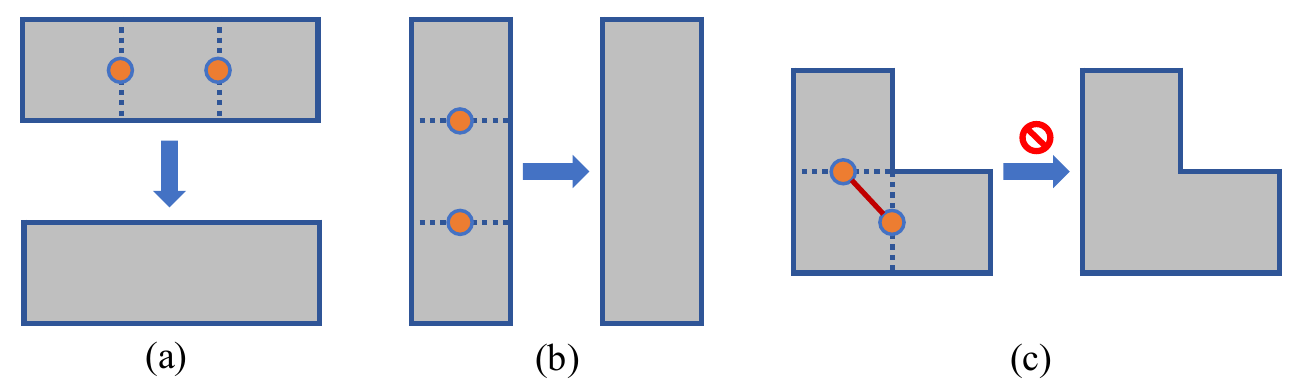}
	\caption{Example of line segments elimination.}
	\label{Fig.Merging}
	\vspace{-1em}
\end{figure}

Let's draw a graph $\mathcal{I}$ that each vertex in $\mathcal{I}$ represents a line segment between two non-occupied grid and there is an undirected edge for every two perpendicular segments that share an endpoint. Deleting two segments that share an endpoint will create a L-shaped block, which is illegal. Illegal merging is shown at Fig. \ref{Fig.Merging}-(c), while legal ones are shown at Fig. \ref{Fig.Merging}-(a)$\sim$(b). The blue dashed lines in Fig. \ref{Fig.Merging} indicate line segments between two non-occupied grids. And each orange dot on the blue dashed line indicates the vertex corresponding to that line segment in the graph $\mathcal{I}$. The red line in Fig. \ref{Fig.Merging}-(c) shows that we only connect vertices for every two perpendicular segments that share an endpoint. Deleting two segments that share an endpoint is equivalent to choosing two connected vertices in $\mathcal{I}$. Therefore, we can only select vertices that are not connected to each other, and these vertices form an independent set of $\mathcal{I}$. Then there is a one-to-one correspondence between the brick tiling of the spanning graph and the independent set of $\mathcal{I}$. In graph theory, an independent set is a set of vertices that there is no edge connecting any two vertices. If a vertex $v$ of $\mathcal{I}$ corresponds to a line segment $s$ in spanning graph $\mathcal{H}$, then $v$ belongs to a given independent set exactly when the two grids separated by $s$ belong to the same brick as each other in the corresponding tiling.

In this case, let $R$, $S$, $T$ denote the number of bricks, the number of non-occupied grids, and the size of the independent set, respectively. Correspondingly, we have the equation

\begin{equation}
	R = S - T
\end{equation}

\begin{figure*}[htbp]
	\vspace{1em}
	\centering
	\includegraphics[width=16cm]{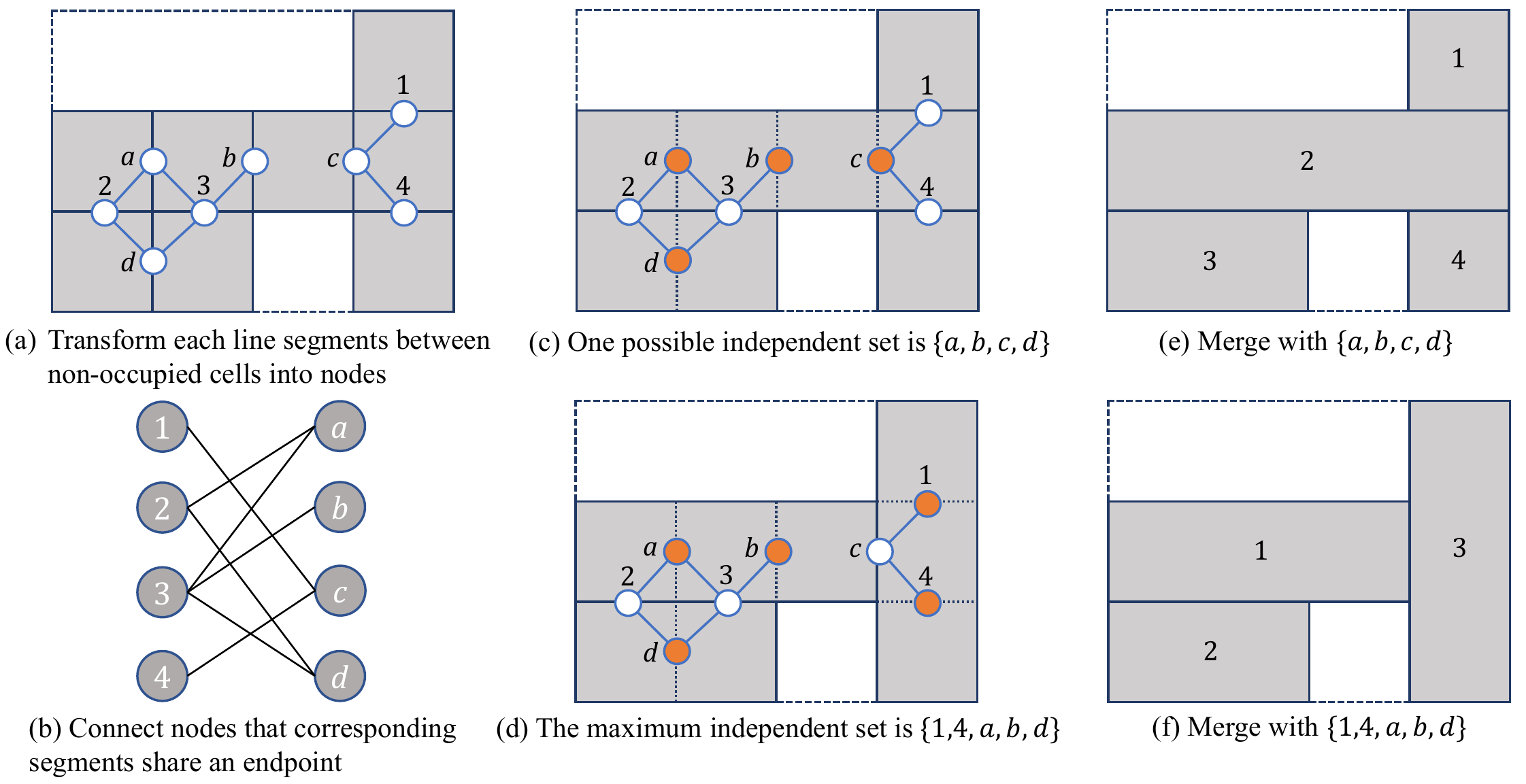}
	\caption{The graph $\mathcal{I}$ and independent vertex set corresponding to 2 different ways of tiling. }
	\label{Fig.NetworkFlow}
	\vspace{-1em}
\end{figure*}

If we do not eliminate any segments at the beginning, we will have $S$ bricks. Each vertex added to the independent set reduces the number of bricks by one. Since $S$ is a fixed value, minimizing $R$ is the same as maximizing $T$. So the minimum brick cover of spanning graph $\mathcal{H}$ corresponds to the maximum independent set of $\mathcal{I}$. The optimization problem of finding such a set is called maximum independent set problem, which is strongly NP-hard. However, $\mathcal{I}$ is a bipartite graph because the vertices corresponding to horizontal segments are only adjacent to the vertices corresponding to vertical segments. It has been proven that the complement of minimum vertex cover is a maximum independent set.  Furthermore, K{\"{o}}nig theorem states that for bipartite graphs, the minimum vertex cover is equivalent to maximum matching, which can be solved in polynomial time using algorithms related to maximum flow\cite{biggs1986graph}. 

Fig. \ref{Fig.NetworkFlow} provides an example to show minimum brick tiling on a $3\times4$ regular grid graph using our algorithm. We first label vertical line segments between free cells with lowercase letters and horizontal ones with numbers in Fig. \ref{Fig.NetworkFlow}-(a). Each line border represents a node in Fig. \ref{Fig.NetworkFlow}-(b) and we put an edge between two nodes when their corresponding segments share the same endpoint to form a bipartite graph $\mathcal{I}$. As shown in Fig. \ref{Fig.NetworkFlow}-(c) and (d), there are two different independent set of $\mathcal{I}$, $\{a,b,c,d\}$ and $\{1,4,a,b,d\}$, which are represented with orange nodes. The latter one is the maximum. Fig. \ref{Fig.NetworkFlow}-(e) and (f) show that if we merge bricks using set $\{a,b,c,d\}$ we will end up with 4 bricks. However, merging with $\{1,4,a,b,d\}$ will only end up with 3 bricks, which is the minimum in this example.

In summary, we transform the minimum brick tiling problem into solving the maximum flow in a flow network. We use Dinic's algorithm to find the vertices of maximum flow in our implementation\cite{dinitz1970algorithm}. Dinic's algorithm runs in $O(\sqrt{V}E)$ time where $V$ and $E$ represent the size of vertices and edges in bipartite graph $\mathcal{I}$. The complement of this set is the maximum independent set. Eventually, we can get the minimum brick tiling of $\mathcal{H}$ by eliminating segments corresponding to vertices in maximum independent set.

\subsection{Combining Bricks Into Spanning Tree}
\begin{figure}[htbp]
	\vspace{0em}
	\centering
	\includegraphics[width=0.35\textwidth]{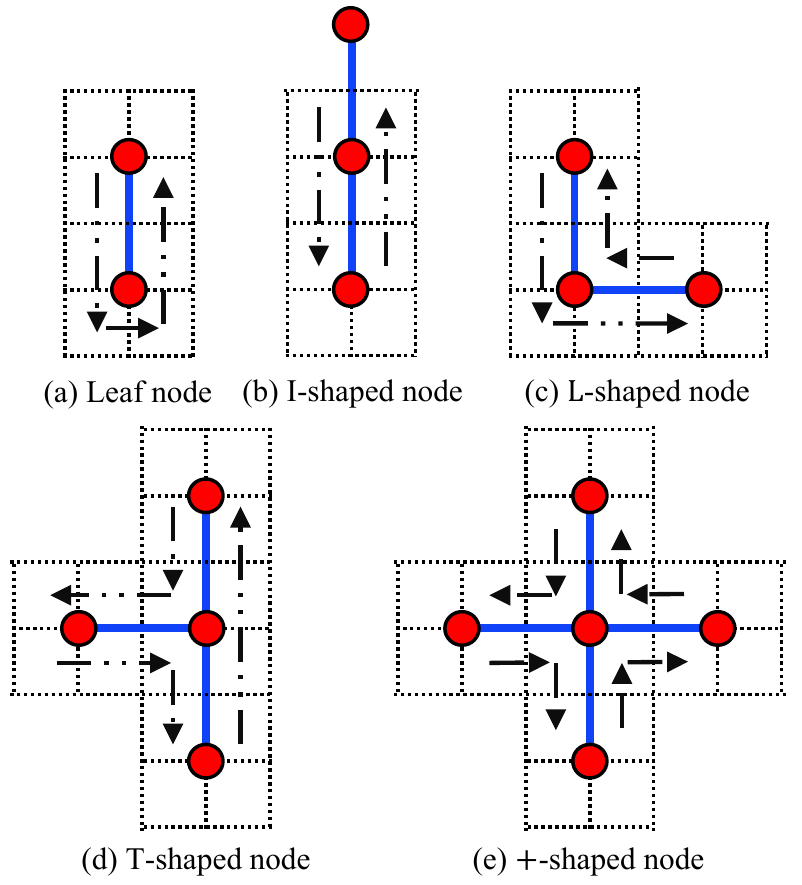}
	\caption{Five different kinds of nodes in spanning graph $\mathcal{H}$. The dotted black arrow indicates the robot's trajectory around the node.}
	\label{Fig.NodeType}
	\vspace{-1em}
\end{figure}

After having least number of bricks as spanning tree branches, we want to merge them for constructing a tree containing all free cells with the rest of unused edges. When robots move along the spanning tree, they produce different number of turns at various kinds of vertex. Fig. \ref{Fig.NodeType} shows five kinds of nodes with various turns. Let function $f(\rho)$ represent the number of turns around tree vertex $\rho$.
\begin{equation}
	f(\rho)=\left\{\begin{array}{ll}
		0 & deg(\rho) = 2 \text{, $\rho$ and its two neighbors}\\
		&  \text{are in the same line} \\
		2 & deg(\rho) = 2 \text{, $\rho$ and its two neighbors}\\
		&  \text{are not in the same line} \\
		2 & deg(\rho) = 1 \text{ or } deg(\rho) = 3 \\
		4 & deg(\rho) = 4
	\end{array}\right.
\end{equation}
where $deg(\rho)$ represents the degree of vertex $\rho$ in the spanning tree. The degree of a vertex is the number of edges that are incident to the vertex.

Each time a new edge is added, the number of degrees and connections of the vertices is also changing, which in turn affects the number of turns. We define function $g(e_{\rho_i, \rho_j})$ as the cost of a new edge. It would be
\begin{equation}
	g(e_{\rho_i, \rho_j}) = f^{+}(\rho_i) + f^{+}(\rho_j) - f(\rho_i) - f(\rho_j)
\end{equation}
where $f^{+}(\rho_i)$, $f^{+}(\rho_i)$ represent turns of $\rho_i$, $\rho_j$ after adding edge $e_{\rho_i, \rho_j}$ respectively.

It's natural to greedily take the least consumed edge at a time and use it to combine bricks so as to construct a spanning tree gradually. Our greedy strategy is summarized as the pseudo-code in Algorithm \ref{alg1}. We calculate the cost of all unused edges and use a binary min-heap to maintain them. As long as there are unmerged bricks, we pop out top elements of min-heap each at a time and check whether it connects vertices from disjoint sets or not. If so, we use such edge to unite two separate bricks unless its cost hasn't changed by now, otherwise we update its cost and insert it into min-heap again. Function \texttt{Union} and \texttt{Find} in Algorithm \ref{alg1} represent merging two sets and finding sets' representatives in union–find data structure, respectively.

\begin{figure}[htbp]
	\centering
	\includegraphics[width=0.49\textwidth]{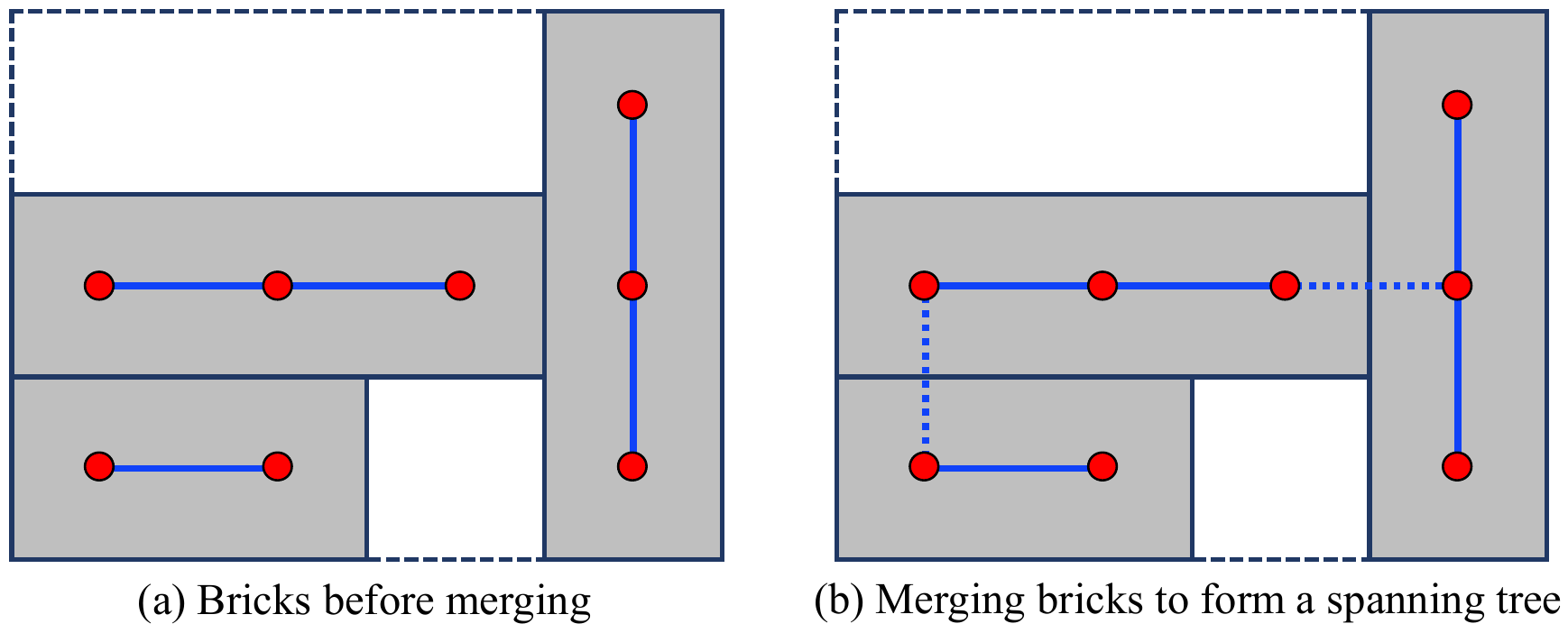}
	\caption{Merging bricks with the rest of edges in spaning graph $\mathcal{H}$. The dotted lines denotes edges connect different bricks. }
	\label{Fig.FormST}
	\vspace{-1em}
\end{figure}

\begin{algorithm}[!htbp]
	\caption{Merge Bricks}
	\label{alg1}
	\LinesNumbered 
	\KwIn {vertex set of bricks $\{\mathnormal{B_1, B_2, ..., B_m}$\} and empty min-heap $H$ of edges}
	\KwOut{edge set $\mathnormal{T}$ of spanning tree in graph $\mathcal{H}$}
	$\mathnormal{T}$ = $\varnothing$ \;
	\For{$i\leftarrow 1$ \KwTo $m$}{
		\For{$j\leftarrow 1$ \KwTo $\left\| B_i \right\| - 1$}{
			\texttt{Union}($\rho_{j}^{i}$, $\rho_{j+1}^{i}$) \;
			$\mathnormal{T}$ = $\mathnormal{T}$ $\cup$ $e_{\rho_{j}^{i}, \rho_{j+1}^{i}}$ \;
		}
	}
	\ForEach{$e_{{\rho_i},{\rho_j}}\in\mathcal{H}$}{
		\If{$e_{{\rho_i},{\rho_j}}\notin\mathnormal{T}$}{
			$e_{{\rho_i},{\rho_j}}.cost = $ \texttt{ComputeEdgeCost}($e_{{\rho_i},{\rho_j}}$) \;
			$\mathnormal{H}.insert(e_{{\rho_i},{\rho_j}})$ \;
		}
	}
	\While{$\mathnormal{H}$ is not empty}{
		$e_{{\rho_i},{\rho_j}}$ = $\mathnormal{H}.top()$ \;
		$\mathnormal{H}.pop()$ \;
		\If{\rm\texttt{Find}$(\rho_i)$  == \rm\texttt{Find}$(\rho_j)$}{
			\bf{continue} \;
		}
		$c$ = \texttt{ComputeEdgeCost$(e_{{\rho_i},{\rho_j}})$} \;
		\eIf{$e_{{\rho_i},{\rho_j}}.cost$ == $c$}{
			$\mathnormal{T}$ = $\mathnormal{T}$ $\cup$ $e_{{\rho_i},{\rho_j}}$ \; 
			\texttt{Union}$(\rho_i, \rho_j)$ \;
		}
		{
			$e_{{\rho_i},{\rho_j}}.cost$ = $c$ \;
			$\mathnormal{H}.insert$($e_{{\rho_i},{\rho_j}}$) \;
		}
		
	}
	\Return{$T$}
\end{algorithm}

\subsection{Improved MSTC* With Turning Cost}
\begin{figure}[!htbp]
	\centering
	\includegraphics[width=0.4\textwidth]{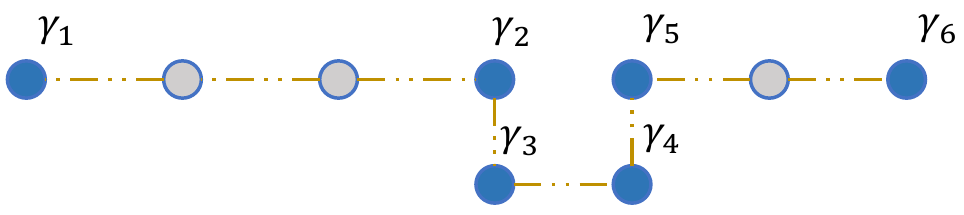}
	\caption{Example of twists(painted in blue) in a covering path. Let the robot starts from $\gamma_1$ and stops at $\gamma_6$. The robot makes four turns during the coverage.}
	\label{Fig.Twist}
\end{figure}

The original MSTC* algorithm does not take into account the robot turning cost, which is equivalent to assuming that the robot only performs uniform motion while covering. In fact, rotation slows the robot down and increases energy consumption. In order to balance the robot's coverage time in practice, in addition to the path length, we need to incorporate the cost of turns into robot's accumulating cost $\mathcal{W}_{\Pi_i}$. Let the acceleration, maximum linear velocity and angular velocity of the robot be $a$, $v_{max}$ and $\omega$. Here, we use $\gamma$ to represent a twist where robot stops and makes a $90^{\circ}$ turn. Each robot's coverage path ${\Pi}_i$ has a twist set ${\Gamma}_i$ corresponding to it.
\begin{equation}
	\Gamma_i = \{ \gamma_1, \gamma_2, \cdots, \gamma_n \}
\end{equation}
$\gamma_1$ and $\gamma_n$ denote start and end points of the coverage path $\Pi_i$, respectively. Fig. \ref{Fig.Twist} shows an example of twists in a robot covering path.

To simplify the problem, we assume that robots perform uniformly accelerated motion and stop at every twist so as to rotate during the whole coverage task. Based on the path, we can calculate the amount of time $T_i$ each robot needs to complete covering the grids with the following formula. 
\begin{normalsize}
\begin{equation}
\begin{array}{cc}
	T_i = \sum\limits_{j=1}^{n-1}t_j + \frac{(n-2)\pi}{4\omega} \vspace{1ex}\\
	t_j=\left\{\begin{array}{ll}
		\sqrt{\frac{2d_{\gamma_j, \gamma_{j+1}}}{a}} 		& d_{\gamma_j, \gamma_{j+1}}\le \frac{v_{max}^{2}}{2a} \vspace{1ex} \\
		\frac{d_{\gamma_j, \gamma_{j+1}}}{v_{max}} + \frac{v_{max}}{2a}  & \mathrm{otherwise}
	\end{array}\right.
\end{array}
\end{equation}
\end{normalsize}where $d_{\gamma_j, \gamma_{j+1}}$ and $t_j$ denote euclidean distance and covering time between $\gamma_j$ and $\gamma_{j+1}$. We use covering time $T_i$ as the cost of the path $\mathcal{W}_{\Pi_i}$ in our implementation.

\section{Results \& Analysis}
In this section, we conduct experiments to test the performance of TMSTC* and compare it to MSTC*, \cite{vandermeulen2019turn} and DARP. We also compare our algorithm to three commonly used methods for constructing spanning tree——DFS(depth first search), ACO(ant colony optimization) and Kruskal's algorithm. The experimental validation focuses on demonstrating the performance differences of candidate algorithms in terms of running time and the number of turns under different environments.

We implement candidate algorithms compiled by C++ on ROS Melodic and use Gazebo with TurtleBot kobuki model for physics simulation. Table \ref{Tab.RobotParams} shows the kobuki kinematic parameters. We have created a regular indoor map and an outdoor map with random obstacles as benchmark environments. In terms of the performance of large-scale mCPP problems, we additionally test candidate algorithms in two street maps NewYork\_0 and Denver\_2, which are discretized from building/road maps from Real-World Benchmarks\cite{sturtevant2012benchmarks}. These environments appear in Fig. \ref{Fig.ExperimentResult}-(a)$\sim$(d), where dark areas indicate obstacles and white areas the space to cover. 

\begin{table}[!htbp]
	\vspace{0em}
	\centering
	\caption{Robot parameters used for simulations}
	\label{Tab.RobotParams}
	\begin{tabular}{l|l}
		\toprule[1pt]
		Coverage Tool Width			   & $0.5m$      \\
		Maximum Translational Velocity & $0.5m/s$    \\
		Maximum Rotational Velocity    & $0.8rad/s$  \\
		Translational Acceleration     & $0.6m/s^{2}$  \\
		\bottomrule[1pt]
	\end{tabular}
	\vspace{0em}
\end{table}

\subsection{Spanning Trees Comparisons}
We test the four candidate algorithms for constructing spanning tree on the four benchmarks environments, and Table \ref{Tab.Turns} reports the number of turns of single robot spanning tree coverage. ACO’s parameters were set to be the same as the parameters adopted by Gao\cite{gao2018optimal}. ACO has fewer turns than DFS on indoor and random environments with small area, but has more turns on Denver\_2 and NewYork\_0. The reason is that the ACO's optimization involves randomness, so the heuristic of reducing turns is not a robust decision. As the size of the map increases, so does the randomness, which makes it difficult for ACO to converge so as to search for a better solution. DFS performs poorly in narrow regions because its spanning tree is always oriented in a single direction, which leads to a great number of turns. Surprisingly, Kruskal performs better than ACO and DFS in general, although Kruskal does not guarantee a reduction of the number of turns. However, Kruskal still exhibits a lack of flexibility in tight space, creating many turns that could have been avoided. Comparing our algorithm with Kruskal, we can see that it reduces the number of turns by 22.4\% on average due to the effectiveness of minimum brick tiling and greedy merging strategy. Experimental data shows that our algorithm can significantly reduce the number of turns in a variety of environments compared to other spanning tree algorithms.

\begin{table}[!htbp]
	\vspace{0em}
	\centering
	\caption{The number of turns}
	\label{Tab.Turns}
	\renewcommand\arraystretch{1.2}
	\begin{tabular}{c||p{0.8cm}<{\centering}||p{0.8cm}<{\centering}||c||p{0.8cm}<{\centering}} 
		Terrain      & DFS        & ACO & Kruskal  & Ours  \\
		\hline
		\hline
		Indoor       & 296        & 254     & 272  & 206    \\
		Random       & 296        & 288     & 258  & 208    \\
		Denver\_2  & 1000       & 1076    & 894  & 652   \\
		NewYork\_0 & 1180       & 1440    & 954  & 780    \\
		\hline
		\hline
		Total      & 2772 & 3058& 2378& 1846 
	\end{tabular}
	\vspace{-2em}
\end{table}

\subsection{Running Time Comparisons}
The parameters of the TMSTC*, MSTC*, \cite{vandermeulen2019turn} and DARP algorithms in the experiments were set as follows: The number of robots varied from 1 to 5 in indoor and outdoor maps, and from 1 to 16 in NewYork\_0 and Denver\_2. The maximum iterations of heuristic method for finding the minimum coverage lines in \cite{vandermeulen2019turn} was 5000. To make DARP iterate sufficiently, the maximum number of iterations of the DARP was set to be 50000 and robots were distributed evenly in the configuration space. DARP adapts DFS for constructing spanning tree while MSTC* uses Kruskal, just as their authors did. 

The results obtained from simulation experiments are presented in Table \ref{Tab.ExpRes}. The total time for a coverage task consists of coverage time and planning time. In Table \ref{Tab.ExpRes}, [min] and [max] represent the time spent by the first and the last robot to complete coverage under different experimental conditions. The difference between the [max] and [min] gives an indication of how balanced the coverage time of the robots are. 

\begin{figure*}[!htbp]
	\vspace{4pt}
	\centering
	\includegraphics[width=18cm]{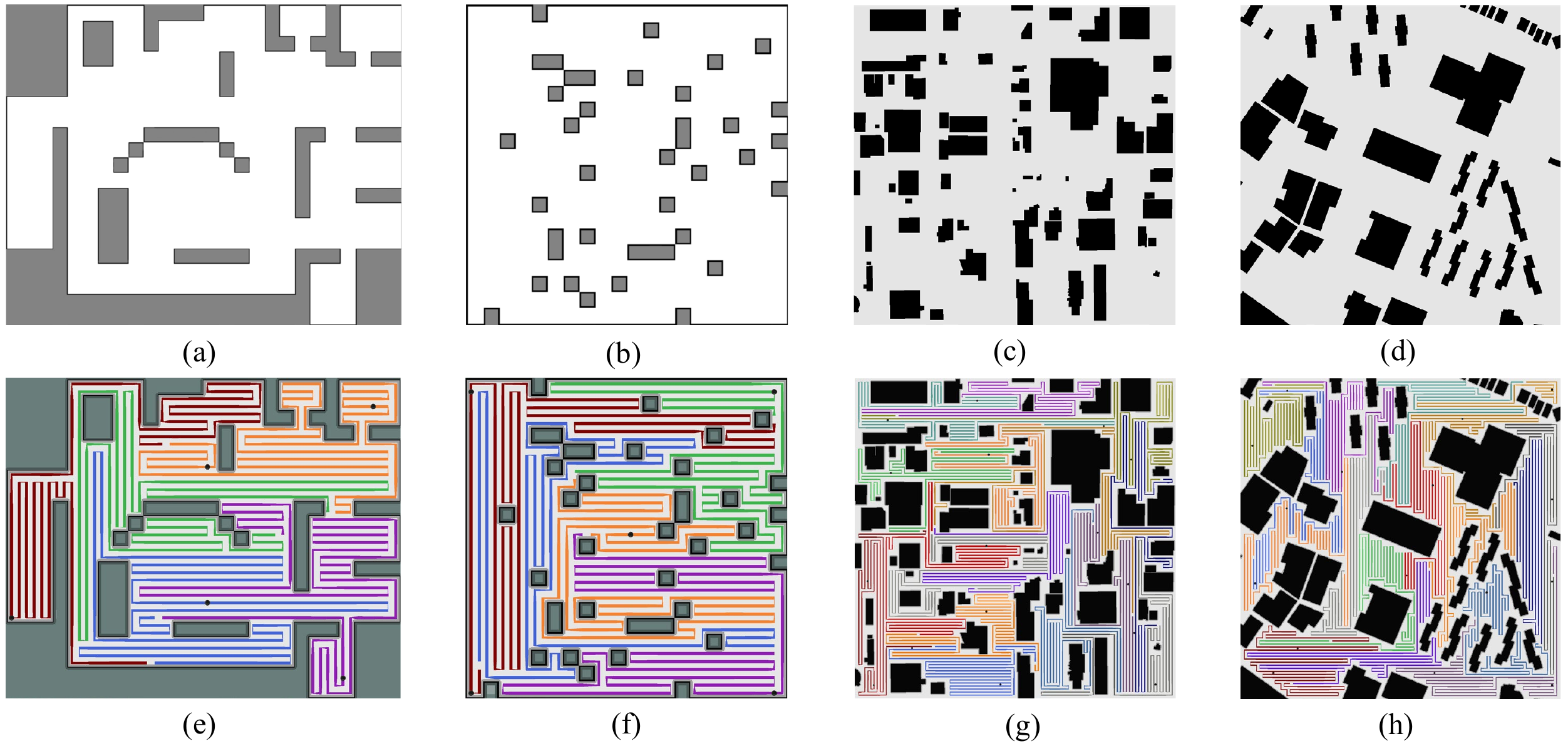}
	\caption{Four different environments. (a) Regular indoor map with ${26\times21}$ mega cells; (b) Outdoor map with ${20\times20}$ mega cells and 10\% obstacles; (c) Denver\_2 with ${54\times54}$ mega cells; (d) NewYork\_0 with ${54\times54}$ mega cells; (e)$\sim$(h) The corresponding results of coverage path planning using our TMSTC* algorithm. We apply 5 robots in (e)(f) and 16 robots in (g)(h).}
	\label{Fig.ExperimentResult}
\end{figure*}

Comparing the coverage time of the four algorithms, we find that TMSTC* outperforms MSTC*, \cite{vandermeulen2019turn} and DARP in every scenario. Fig. \ref{Fig.ExperimentResult}-(e)$\sim$(h) show the planning path of TMSTC* appearing on rviz. The difference between [max] and [min] is also smaller than the other three algorithms because TMSTC* takes into account the turning time when assigning paths to multiple robots. The total coverage time of our algorithm in four different maps is 5.9\% smaller than that of \cite{vandermeulen2019turn} and 8.3\% smaller than that of MSTC*. Although the reduction of turns results in smaller coverage time, the approach of \cite{vandermeulen2019turn} uses m-TSP to assign coverage lines to each robot instead of grids, and m-TSP does not incorporate turns into the cost, which in turn leads to an imbalance in robots' coverage time. The large area maps Denver\_2 and NewYork\_0 exacerbate this imbalance. DARP divides continuous areas into similar size, however, the shape of these areas are completely random and DARP dose not take the number of turns into consideration, which directly leads to the inability of DARP to equalize the coverage time of the robots and thus increases the total time spent on the coverage task. 


The planning time of TMSTC* is very similar to that of MSTC* for the same scenario and the maximum difference between the two is no more than 5s. Overall, the planning time of TMSTC* is even smaller than that of MSTC*. DARP plans significantly faster than TMSTC*, MSTC* and \cite{vandermeulen2019turn} in small maps, but far slower than the latter two in large areas. DARP uses gradient descent to allocate areas of equal size for each robot, which is difficult to converge in large maps. In summary, TMSTC* has a smaller planning time than that of MSTC*, \cite{vandermeulen2019turn}, and DARP.

\begin{table*}
	\vspace{0em}
	\centering
	\caption{Experimental results of four different algorithms}
	\label{Tab.ExpRes}
	\renewcommand\arraystretch{1.15}
	\scalebox{0.9}{
	\begin{tabular}{c|c||cc|c||cc|c||cc|c||cc|c}
		\multirow{3}{*}{Map}    & \multirow{3}{*}{Robots} & \multicolumn{3}{c||}{TMSTC*} & \multicolumn{3}{c||}{MSTC*}  & \multicolumn{3}{c||}{\cite{vandermeulen2019turn}} & \multicolumn{3}{c}{DARP}                                               \\
		&                         & \multicolumn{2}{c|}{Coverage Time (s)} & \multirow{2}{1cm}{\centering Planning Time (s)} & \multicolumn{2}{c|}{Coverage Time (s)} & \multirow{2}{1cm}{\centering Planning Time (s)} & \multicolumn{2}{c|}{Coverage Time (s)} & \multirow{2}{1cm}{\centering Planning Time (s)} & \multicolumn{2}{c|}{Coverage Time (s)} & \multirow{2}{1cm}{\centering Planning Time (s)}  \\
		
		&                         & Max     & (Min)                   &                                   & Max     & (Min)                   &                                   & Max     & (Min)                   &                                   & Max     & (Min)                   &                                    \\
		\hline
		\hline
		\multirow{5}{*}{Indoor}     & 1                       & \textbf{2020.3}  & \textbf{2020.3}                  & 0.005                             & 2178.6  & 2178.6                  & 0.001                             & 2143.2  & 2143.2                  & 0.519                             & 2274.3  & 2274.3                  & 0.002                              \\
		& 2                       & \textbf{1101.4}  & \textbf{1068.5}                  & 0.416                             & 1184.2  & 1151.2                  & 0.467                             & 1156.9  & 1122.2                  & 0.753                             & 1194.9  & 1190.1                  & 0.004                              \\
		& 3                       & \textbf{738.4}   & \textbf{719.5}                   & 2.134                             & 795.8   & 774.9                   & 1.203                             & 789.4   & 745.4                   & 1.029                             & 837.0   & 772.9                   & 0.010                              \\
		& 4                       & \textbf{576.7}   & \textbf{538.7}                   & 2.121                             & 629.3   & 556.2                   & 1.173                             & 596.1   & 566.5                   & 1.563                             & 651.7   & 553.2                   & 0.034                              \\
		& 5                       & \textbf{466.8}   & \textbf{449.4}                   & 3.420                             & 516.5   & 455.6                   & 1.992                             & 504.6   & 456.5                   & 2.383                             & 554.4   & 452.4                   & 0.065                              \\
		\hline
		\multirow{5}{*}{Random}     & 1                       & \textbf{1940.9}  & \textbf{1940.9}                  & 0.006                             & 2049.6  & 2049.6                  & 0.001                             & 1991.8  & 1991.8                  & 0.928                             & 2175.7  & 2175.7                  & 0.000                              \\
		& 2                       & \textbf{1025.6}  & \textbf{1018.0}                  & 0.320                             & 1117.2  & 1059.7                  & 0.329                             & 1058.4  & 1023.0                  & 1.264                             & 1161.6  & 1155.6                  & 0.136                              \\
		& 3                       & \textbf{724.9}   & \textbf{699.7}                   & 1.343                             & 804.3   & 707.7                   & 0.668                             & 767.8   & 738.9                   & 2.469                             & 813.5   & 728.2                   & 0.000                              \\
		& 4                       & \textbf{542.2}   & \textbf{525.6}                   & 0.978                             & 633.8   & 528.6                   & 1.993                             & 573.8   & 535.8                   & 2.837                             & 612.5   & 583.3                   & 0.025                              \\
		& 5                       & \textbf{456.9}   & 433.5                   & 2.555                             & 511.0   & 425.8                   & 4.117                             & 476.1   & 386.4                   & 2.275                             & 498.0   & 452.3                   & 0.010                              \\
		\hline
		\multirow{5}{*}{Denver\_2}  & 1                       & \textbf{8027.9}  & \textbf{8027.9}                  & 0.011                             & 8851.7  & 8851.7                  & 0.003                             & 8571.5  & 8571.5                  & 5.771                             & 9012.9  & 9012.9                  & 0.004                              \\
		& 4                       & \textbf{2154.2}  & \textbf{2109.0}                  & \textbf{4.445}                             & 2447.6  & 2228.1                  & 7.862                             & 2294.5  & 2198.7                  & 9.107                             & 2538.2  & 2331.2                  & 253.606                            \\
		& 8                       & \textbf{1129.2}  & \textbf{1066.0}                  & 16.184                            & 1318.4  & 1132.8                  & 15.061                            & 1191.2  & 1099.1                  & 13.862                            & 1356.3  & 1267.6                  & 310.838                            \\
		& 12                      & \textbf{790.2}   & 744.2                   & 20.044                            & 904.9   & 762.0                   & 17.365                            & 837.6   & 719.6                   & 19.771                            & 931.8   & 801.7                   & 249.274                            \\
		& 16                      & \textbf{606.1}   & 546.0                   & \textbf{13.208}                            & 712.0   & 552.3                   & 15.021                            & 670.3   & 576.5                   & 20.635                            & 705.7   & 518.0                   & 286.738                            \\
		\hline
		\multirow{5}{*}{NewYork\_0} & 1                       & \textbf{8484.9}  & \textbf{8484.9}                  & 0.016                             & 8851.7  & 8851.7                  & 0.004                             & 8862.4  & 8862.4                  & 13.992                            & 9473.9  & 9473.9                  & 0.003                              \\
		& 4                       & \textbf{2234.6}  & \textbf{2187.4}                  & 6.429                             & 2447.6  & 2228.1                  & 5.581                             & 2387.9  & 2282.2                  & 12.740                            & 2559.9  & 2425.4                  & 95.433                             \\
		& 8                       & \textbf{1198.6}  & 1138.3                  & 11.355                            & 1318.4  & 1132.8                  & 9.958                             & 1385.4  & 1193.9                  & 22.837                            & 1397.8  & 1108.8                  & 174.158                            \\
		& 12                      & \textbf{832.7}   & 771.5                   & \textbf{12.674}                            & 904.9   & 762.0                   & 17.365                            & 951.1   & 744.7                   & 20.453                            & 945.0   & 746.9                   & 193.175                            \\
		& 16                      & \textbf{610.4}   & 578.2                   & 16.791                            & 712.0   & 552.3                   & 15.021                            & 676.9   & 557.9                   & 15.085                            & 739.0   & 597.3                   & 236.250                            \\
		\hline
		\hline
		\multicolumn{2}{c||}{Total}  & \textbf{35662.6} & \textbf{35067.4}                 & \textbf{114.455}                           & 38889.7 & 36941.7                 & 115.185                           & 37886.8 & 36516.1                 & 170.273                           & 40434.1 & 38621.9                 & 1799.765                          
	\end{tabular}
	}
\vspace{0em}
\end{table*}

\section{Conclusions \& Future Work}
In this paper, we propose the TMSTC* coverage planning algorithm, whose goal is to minimize the number of turns to reduce the coverage time. Our algorithm improves MSTC* by constructing a tree that allows robots to make fewer turns when circumnavigating, and incorporates turning cost when balancing weights. Comparative experiments with the MSTC*, \cite{vandermeulen2019turn} and DARP algorithms show that TMSTC* has advantages in both simple and complex configuration spaces in terms of planning time.

Fault tolerance is vital for multi-robot frameworks, particularly for those worked in remote environments. For future work, we intend to augment TMSTC* to tolerate failures so as to replan trajectories for the functional robots without immediate human intervention.

\addtolength{\textheight}{-0cm}   




\section*{Acknowledgment}

This work was supported in part by the National Science Foundation of China under Grant 62172111, National Joint Fund Key Project (NSFC- Guangdong Joint Fund) under Grant U21A20478, and in part by Major Science and Technology Projects of Zhongshan under Grant 191018182628219.

\bibliographystyle{unsrt}
\bibliography{path_planning}

\begin{thebibliography}{10}

\bibitem{galceran2013survey}
Enric Galceran and Marc Carreras.
\newblock A survey on coverage path planning for robotics.
\newblock {\em Robotics and Autonomous systems}, 61(12):1258--1276, 2013.

\bibitem{yasutomi1988cleaning}
Fumio Yasutomi, Makoto Yamada, and Kazuyoshi Tsukamoto.
\newblock Cleaning robot control.
\newblock In {\em Proceedings. 1988 IEEE International Conference on Robotics
  and Automation}, pages 1839--1841. IEEE, 1988.

\bibitem{hameed2014intelligent}
Ibrahim~A Hameed.
\newblock Intelligent coverage path planning for agricultural robots and
  autonomous machines on three-dimensional terrain.
\newblock {\em Journal of Intelligent \& Robotic Systems}, 74(3):965--983,
  2014.

\bibitem{acevedo2013distributed}
Jose~Joaquin Acevedo, Bego{\~n}a~C Arrue, Ivan Maza, and Anibal Ollero.
\newblock Distributed approach for coverage and patrolling missions with a team
  of heterogeneous aerial robots under communication constraints.
\newblock {\em International Journal of Advanced Robotic Systems}, 10(1):28,
  2013.

\bibitem{acar2003path}
Ercan~U Acar, Howie Choset, Yangang Zhang, and Mark Schervish.
\newblock Path planning for robotic demining: Robust sensor-based coverage of
  unstructured environments and probabilistic methods.
\newblock {\em The International journal of robotics research},
  22(7-8):441--466, 2003.

\bibitem{almadhoun2019survey}
Randa Almadhoun, Tarek Taha, Lakmal Seneviratne, and Yahya Zweiri.
\newblock A survey on multi-robot coverage path planning for model
  reconstruction and mapping.
\newblock {\em SN Applied Sciences}, 1(8):1--24, 2019.

\bibitem{arkin2000approximation}
Esther~M Arkin, S{\'a}ndor~P Fekete, and Joseph~SB Mitchell.
\newblock Approximation algorithms for lawn mowing and milling.
\newblock {\em Computational Geometry}, 17(1-2):25--50, 2000.

\bibitem{arkin2005optimal}
Esther~M Arkin, Michael~A Bender, Erik~D Demaine, S{\'a}ndor~P Fekete,
  Joseph~SB Mitchell, and Saurabh Sethia.
\newblock Optimal covering tours with turn costs.
\newblock {\em SIAM Journal on Computing}, 35(3):531--566, 2005.

\bibitem{tang2021mstc}
Jingtao Tang, Chun Sun, and Xinyu Zhang.
\newblock Mstc*: Multi-robot coverage path planning under physical constrain.
\newblock In {\em 2021 IEEE International Conference on Robotics and Automation
  (ICRA)}, pages 2518--2524. IEEE, 2021.

\bibitem{vandermeulen2019turn}
Isaac Vandermeulen, Roderich Gro{\ss}, and Andreas Kolling.
\newblock Turn-minimizing multirobot coverage.
\newblock In {\em 2019 International Conference on Robotics and Automation
  (ICRA)}, pages 1014--1020. IEEE, 2019.

\bibitem{gabriely2001spanning}
Yoav Gabriely and Elon Rimon.
\newblock Spanning-tree based coverage of continuous areas by a mobile robot.
\newblock {\em Annals of mathematics and artificial intelligence},
  31(1):77--98, 2001.

\bibitem{gabriely2002spiral}
Yoav Gabriely and Elon Rimon.
\newblock Spiral-stc: An on-line coverage algorithm of grid environments by a
  mobile robot.
\newblock In {\em Proceedings 2002 IEEE International Conference on Robotics
  and Automation (Cat. No. 02CH37292)}, volume~1, pages 954--960. IEEE, 2002.

\bibitem{hazon2005redundancy}
Noam Hazon and Gal~A Kaminka.
\newblock Redundancy, efficiency and robustness in multi-robot coverage.
\newblock In {\em Proceedings of the 2005 IEEE international conference on
  robotics and automation}, pages 735--741. IEEE, 2005.

\bibitem{hazon2006towards}
Noam Hazon, Fabrizio Mieli, and Gal~A Kaminka.
\newblock Towards robust on-line multi-robot coverage.
\newblock In {\em Proceedings 2006 IEEE International Conference on Robotics
  and Automation, 2006. ICRA 2006.}, pages 1710--1715. IEEE, 2006.

\bibitem{agmon2006constructing}
Noa Agmon, Noam Hazon, and Gal~A Kaminka.
\newblock Constructing spanning trees for efficient multi-robot coverage.
\newblock In {\em Proceedings 2006 IEEE International Conference on Robotics
  and Automation, 2006. ICRA 2006.}, pages 1698--1703. IEEE, 2006.

\bibitem{even2004min}
Guy Even, Naveen Garg, Jochen K{\"o}nemann, Ramamoorthi Ravi, and Amitabh
  Sinha.
\newblock Min--max tree covers of graphs.
\newblock {\em Operations Research Letters}, 32(4):309--315, 2004.

\bibitem{zheng2005multi}
Xiaoming Zheng, Sonal Jain, Sven Koenig, and David Kempe.
\newblock Multi-robot forest coverage.
\newblock In {\em 2005 IEEE/RSJ International Conference on Intelligent Robots
  and Systems}, pages 3852--3857. IEEE, 2005.

\bibitem{zheng2007robot}
Xiaoming Zheng and Sven Koenig.
\newblock Robot coverage of terrain with non-uniform traversability.
\newblock In {\em 2007 IEEE/RSJ International Conference on Intelligent Robots
  and Systems}, pages 3757--3764. IEEE, 2007.

\bibitem{zheng2010multirobot}
Xiaoming Zheng, Sven Koenig, David Kempe, and Sonal Jain.
\newblock Multirobot forest coverage for weighted and unweighted terrain.
\newblock {\em IEEE Transactions on Robotics}, 26(6):1018--1031, 2010.

\bibitem{sun2021ft}
Chun Sun, Jingtao Tang, and Xinyu Zhang.
\newblock Ft-mstc*: An efficient fault tolerance algorithm for multi-robot
  coverage path planning.
\newblock In {\em 2021 IEEE International Conference on Real-time Computing and
  Robotics (RCAR)}, pages 107--112. IEEE, 2021.

\bibitem{kapoutsis2017darp}
Athanasios~Ch Kapoutsis, Savvas~A Chatzichristofis, and Elias~B Kosmatopoulos.
\newblock Darp: divide areas algorithm for optimal multi-robot coverage path
  planning.
\newblock {\em Journal of Intelligent \& Robotic Systems}, 86(3):663--680,
  2017.

\bibitem{apostolidis2022cooperative}
Savvas~D Apostolidis, Pavlos~Ch Kapoutsis, Athanasios~Ch Kapoutsis, and Elias~B
  Kosmatopoulos.
\newblock Cooperative multi-uav coverage mission planning platform for remote
  sensing applications.
\newblock {\em Autonomous Robots}, 46(2):373--400, 2022.

\bibitem{gao2018optimal}
Chunqing Gao, Yingxin Kou, Zhanwu Li, An~Xu, You Li, and Yizhe Chang.
\newblock Optimal multirobot coverage path planning: ideal-shaped spanning
  tree.
\newblock {\em Mathematical Problems in Engineering}, 2018, 2018.

\bibitem{ramesh2022optimal}
Megnath Ramesh, Frank Imeson, Baris Fidan, and Stephen~L Smith.
\newblock Optimal partitioning of non-convex environments for minimum turn
  coverage planning.
\newblock {\em IEEE Robotics and Automation Letters}, 7(4):9731--9738, 2022.

\bibitem{biggs1986graph}
Norman Biggs, E~Keith Lloyd, and Robin~J Wilson.
\newblock {\em Graph Theory, 1736-1936}.
\newblock Oxford University Press, 1986.

\bibitem{dinitz1970algorithm}
Yefim~A Dinitz.
\newblock An algorithm for the solution of the problem of maximal flow in a
  network with power estimation.
\newblock In {\em Doklady Akademii nauk}, volume 194, pages 754--757. Russian
  Academy of Sciences, 1970.

\bibitem{sturtevant2012benchmarks}
N.~Sturtevant.
\newblock Benchmarks for grid-based pathfinding.
\newblock {\em Transactions on Computational Intelligence and AI in Games},
  4(2):144 -- 148, 2012.

\end{thebibliography}

\end{document}